\documentclass{article}

\usepackage[svgnames]{xcolor}

\usepackage{microtype}
\usepackage{graphicx}
\usepackage{subfigure}
\usepackage{booktabs} 
\usepackage{hyperref}

\usepackage[accepted]{icml2025}

\usepackage{amsmath}
\usepackage{amssymb}
\usepackage{mathtools}
\usepackage{amsthm}

\usepackage[capitalize,noabbrev]{cleveref}

\theoremstyle{plain}

\theoremstyle{definition}

\theoremstyle{remark}

\usepackage[textsize=tiny]{todonotes}

\newcommand{\methodname}{ET-VLA}

\begin{document}

\twocolumn[
\icmltitle{Embodiment Transfer Learning for Vision-Language-Action Models}




\begin{icmlauthorlist}
\icmlauthor{Chengmeng Li\textsuperscript{1}}{}
\icmlauthor{Yaxin Peng\textsuperscript{1, $\dagger$}}{}

\end{icmlauthorlist}

\icmlaffiliation{shu}{Shanghai University}


\centerline{\textsuperscript{1}Shanghai University}
\vskip 0.5em
]



\printAffiliationsAndNotice{$\dagger$ Corresponding author} 

\begin{figure*}[t]
    \centering
    \vspace{-0.3cm}
    \includegraphics[width=0.95\textwidth]{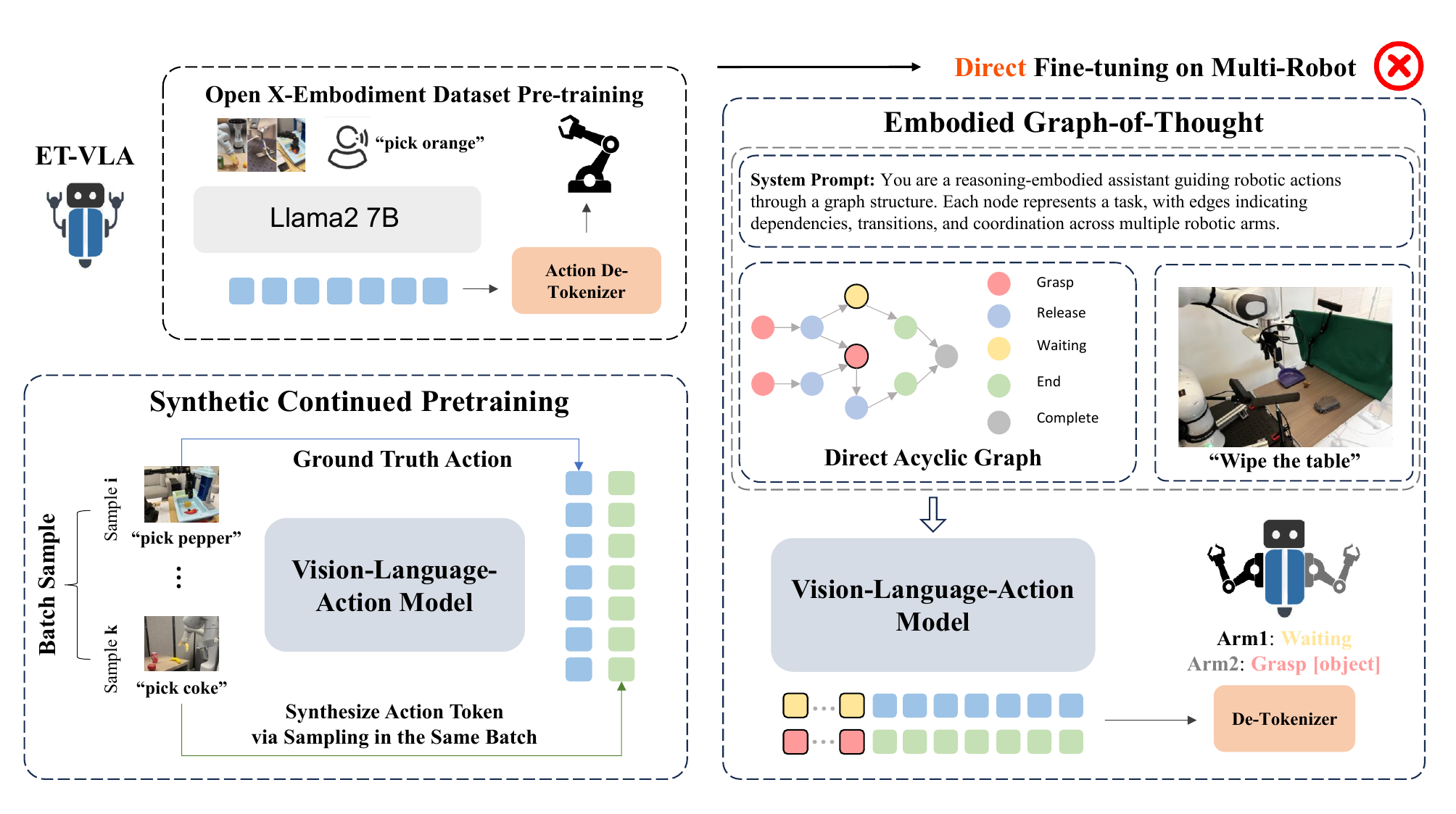}
    \vspace{-0.1cm}
    \caption{\textbf{Overview of \methodname:} we propose \textbf{Synthetic Continued Pretraining (SCP)} and \textbf{Embodied Graph-of-Thought (EGoT)} as key components: \textbf{(1) SCP} uses synthetically generated data to warm up the model for the new embodiment, bypassing the need for real human demonstrations and reducing data collection costs. SCP enables the model to learn correct actions and precise action token numbers. \textbf{(2) Embodied GoT} formulates each sub-task as a node, that allows the VLA model to distinguish the functionalities and roles of each embodiment during task execution, promoting more effective coordination.}
\end{figure*}

\begin{abstract}
Vision-language-action (VLA) models have significantly advanced robotic learning, enabling training on large-scale, cross-embodiment data and fine-tuning for specific robots. However, state-of-the-art autoregressive VLAs struggle with multi-robot collaboration. We introduce embodiment transfer learning, denoted as \textbf{\methodname}, a novel framework for efficient and effective transfer of pre-trained VLAs to multi-robot. \methodname's core is Synthetic Continued Pretraining (SCP), which uses synthetically generated data to warm up the model for the new embodiment, bypassing the need for real human demonstrations and reducing data collection costs. SCP enables the model to learn correct actions and precise action token numbers. Following SCP, the model is fine-tuned on target embodiment data. To further enhance the model performance on multi-embodiment, we present the Embodied Graph-of-Thought technique, a novel approach that formulates each sub-task as a node, that allows the VLA model to distinguish the functionalities and roles of each embodiment during task execution. Our work considers bimanual robots, a simple version of multi-robot to verify our approaches. We validate the effectiveness of our method on both simulation benchmarks and real robots covering three different bimanual embodiments. In particular, our proposed \methodname \space can outperform OpenVLA on six real-world tasks over 53.2\%. We will open-source all codes to support the community in advancing VLA models for robot learning. The project is available at \href{https://et-vla.github.io}{https://et-vla.github.io}
\end{abstract}
\section{Introduction}
\label{sec:intro}
Autoregressive models, which generate sequences of discrete tokens by iteratively predicting the next element conditioned on prior outputs, serve as the foundation for modern large language models. This sequential prediction paradigm has revolutionized robotics, enabling Vision-Language-Action (VLA) models such as RT-2~\cite{brohan2023rt-2} and OpenVLA~\cite{openvla} to map sensory inputs to robotic actions through discretized token predictions. While these models excel in unimanual manipulation tasks, their performance degrades significantly when applied to multi-robot systems. Sometimes they even fail to produce structurally valid action sequences (e.g., generating incorrect numbers of tokens). This raises two critical questions: (1) Why do current VLA models struggle to adapt to multi-robots? (2) How can we enhance their robustness and generalization for deployment on novel multi-robot robotic platforms?

Regarding the first question, we identify a dependency of VLAs on pretraining data. These models rely on vast datasets to learn precise token-to-action mappings, yet existing open-source robotics datasets (e.g., OXE~\cite{o2023open-x}, Droid~\cite{khazatsky2024droid} focus exclusively on unimanual robot systems. Consequently, VLAs encounter a distribution shift when applied to multi-robots, lacking the necessary priors to interpret action tokens for unseen embodiments. To address these limitations, we propose Synthetic Continued Pretraining (SCP). In particular, SCP generates synthetic multi-robot data, enabling cost-effective embodied transfer learning without reliance on expensive teleoperated datasets. Rather than directly learning the mapping between observations and robot actions, SCP focuses on establishing the mapping between the embodiment and action sequence. To further enhance VLA's performance on multi-robot, we propose Embodied Graph-of-Thought (EGoT). It creates an actionable graph that meticulously allocates tasks to each robot, ensuring they act in accordance with the designated task sequence. By combining these two techniques, our proposed \methodname \space can do multi-tasking on a multi-robot with quick fine-tuning.

To assess the effectiveness of our approach, we conducted both simulations and real-world experiments, using bimanual robots, a simple version of multi-robots as examples. Our experiments are conducted on three bimanual robots, including bimanual UR5e, bimanual Franka, and bimanual AgileX. These experiments clearly show that directly using OpenVLA in multi-robot scenarios leads to unacceptably low success rates. In contrast, our ET-VLA model, which leverages OpenVLA's architecture and pre-trained weights, demonstrates a substantial improvement after fine-tuning for specific dexterous tasks. For instance, when tested on six tasks involving two UR5e robots, ET-VLA achieved success rates nine times higher than OpenVLA. These results strongly validate the advantages of our ET-VLA model for multi-robot applications.

In summary, our contributions are threefold:
\begin{itemize}
    \item We provide an in-depth analysis of the limitations of existing autoregressive VLA models in multi-robot multi-task settings.
    \item We propose \methodname, a novel framework incorporating two advanced techniques: Synthetic Continued Pretraining and Embodied Graph-of-Thought, which significantly enhance the performance of VLAs in multi-robot manipulation.
    \item We extensively evaluate \methodname\space on the real robot and simulation, demonstrating its superior performance compared to state-of-the-art VLA models. 
\end{itemize}

\section{Related Works}
\label{sec:related_works}

\textbf{Vision-language-action models.} Vision-language-action models (VLAs) represent a family of work that leverages the power of pre-trained vision-language models as a backbone to understand language as well as observation. Such a method directly fine-tuned large pretrained VLMs~\cite{liu2025mm, zhu2024llava, zhu2024comprehensive, lu2024deepseek-vl, openflamingo, idefics, llava, llava1.5, wang2024qwen2, chen2024internvl, zhu2021unilog, jiang2024ragraph, meng2023logsummary, zhu2022teach, zhou2023make, huang2022label} for predicting robot actions~\cite{brohan2023rt-2, openvla, embodiedcot, roboflamingo, leo3d, wen2024tinyvla, [pi0, pertsch2025fast, wu2025discrete}. Existing works can be largely categorized into two different methods, the one uses next-token prediction as the training objective to generate action tokens autoregressively, similar to the ways language models generate text. The other use policy head, i.e., RNN head~\cite{roboflamingo} or diffusion head~\cite{diffusion-policy, zhu2024scalingdp, wang2024sparse-dp, prasad2024consistencypolicy, black2023training, black2023zero, dasari2024ingredients, lin2024datascalinglawsimitation, multimodal_diffusion_transformer, aloha_unleashed, uehara2024fine, uehara2024feedback, zhu2024language, zhu2024retrieval, wen2024object, li2025pointvla, zhu2025scaling, li2025coa, li2024mmro, zhou2025vision, zhu2025objectvla} to predict robot action. These methods have demonstrated strong performance in both simulations and real-world tasks. However, most are focused on unimanual robot settings. In this work, we show that popular VLAs can fail in multi-robot manipulation tasks. We explore ways to extend these methods to multi-robot without requiring costly re-training.

\begin{figure*}[t]
\centering
    \includegraphics[width=0.95\textwidth]{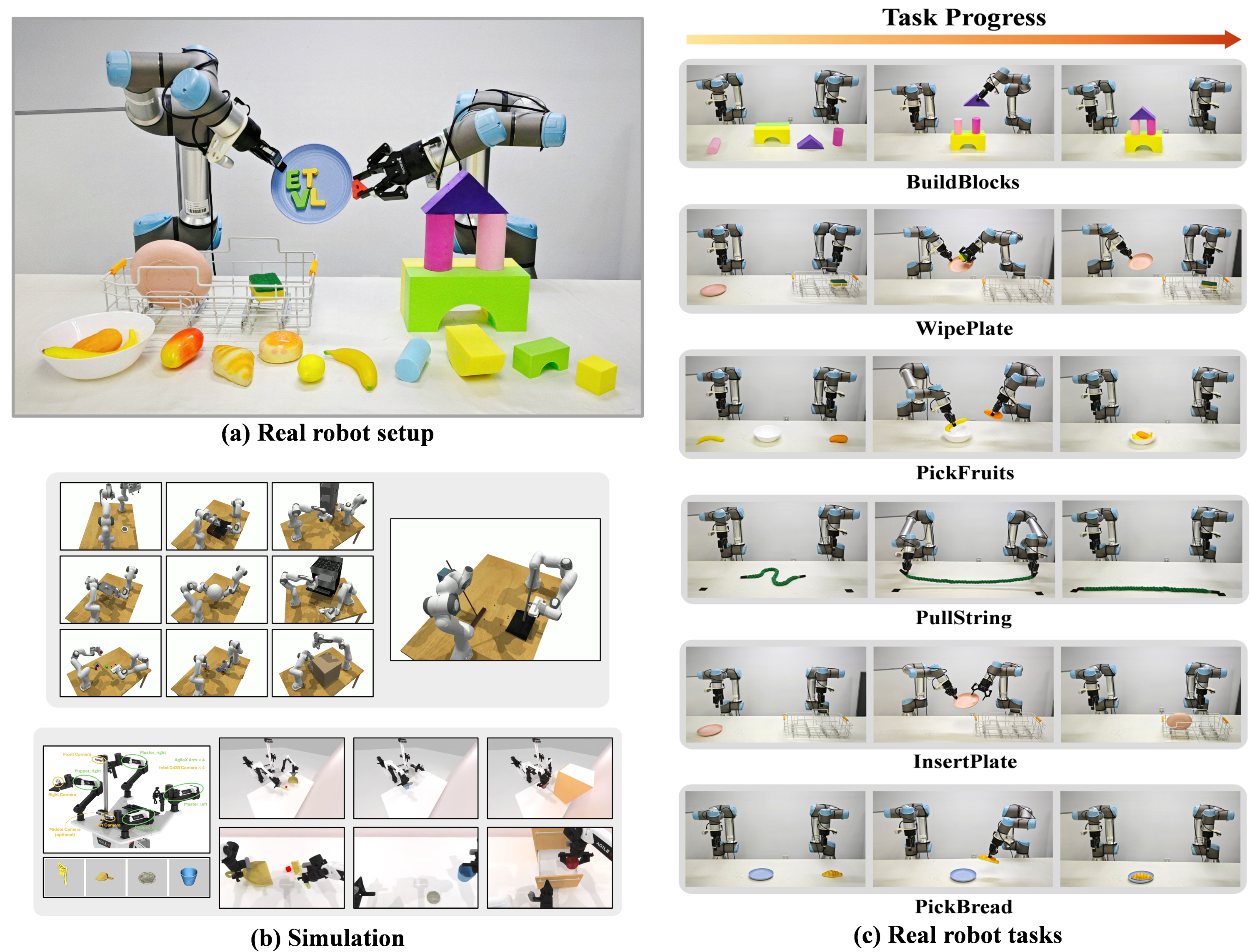}
    \vspace{-0.4cm}
    \caption{\textbf{(a) Real robot and all objects used in our work.} We use two UR5 robot arms equipped with Robotiq grippers and incorporate a diverse set of everyday objects in our manipulation tasks. A RealSense D457 camera is applied to capture visual observations on the top. \textbf{(b) Simulation.} We conduct experiments on two simulation benchmarks, the RLBench2~\cite{grotz2024peract2} and RoboTwin~\cite{mu2024robotwin}. \textbf{(c) Real robot tasks.} We designed six collaborative multi-robot tasks for our real-world experiment.}
    \label{fig:workspace}
    \vspace{-0.40cm} 
\end{figure*}

\textbf{Reasoning in robot learning.} A number of works have explored the use of Large Language Models and Vision-Language Models~\cite{llava1.5} in robotics, particularly for visual representations, object detection~\cite{jiang2022vima, oci}, high-level planning~\cite{rana2023sayplan, chowdhery2023palm, slowfast, niu2024llarva}, and code generation~\cite{codeaspolicy}. ECoT proposes a method to predict an object's position, task plans, and current action to enhance action generation. DAG~\cite{gao2024dag} leverage directed acyclic graph to help LLM perform better planning for multi-robot.  CoT-VLA~\cite{cot-vla} generates sub-goa for autoregressive VLA to guide action. In this work, our proposed Embodied Graph-of-Thought enables the model to differentiate tasks for two robots, facilitating collaboration between the two embodiments.

\section{Methodology}
\subsection{Preliminary: Vision-Language-Action Models}
The architecture of recent Vision-Language-Action (VLA) Models consists of two main components: 1) visual encoders, which map image inputs to feature embeddings and use a projector — typically a few layers of multi-layer perception — to align with the input space of a language model, and 2) a large language model (LLM) backbone that processes all input representations. The model undergoes two primary training phases: a pre-training stage and a fine-tuning stage. During both stages, the model is trained end-to-end using a next-token prediction objective. At inference time, the model accepts both visual observations and language instructions, producing discrete action tokens that control the robot.

This pipeline is widely adopted by various projects, including OpenVLA~\cite{openvla}, one of the most influential and open-source VLA frameworks. OpenVLA is pre-trained on the Open X-Embodiment (OXE in short)~\cite{o2023open} datasets and fine-tuned on the Bridge Data V2~\cite{walke2023bridgedata}, utilizing the Prismatic-7B VLM~\cite{karamcheti2024prismatic} along with 600 million visual encoder parameters. During testing, OpenVLA generates seven action tokens to control the robot, demonstrating strong performance on standard tasks as well as notable generalization capabilities. This work follows the OpenVLA paradigm, addressing the problem and presenting the solution within this framework.

\begin{figure}[t]
\centering
  \includegraphics[width=0.85\columnwidth]{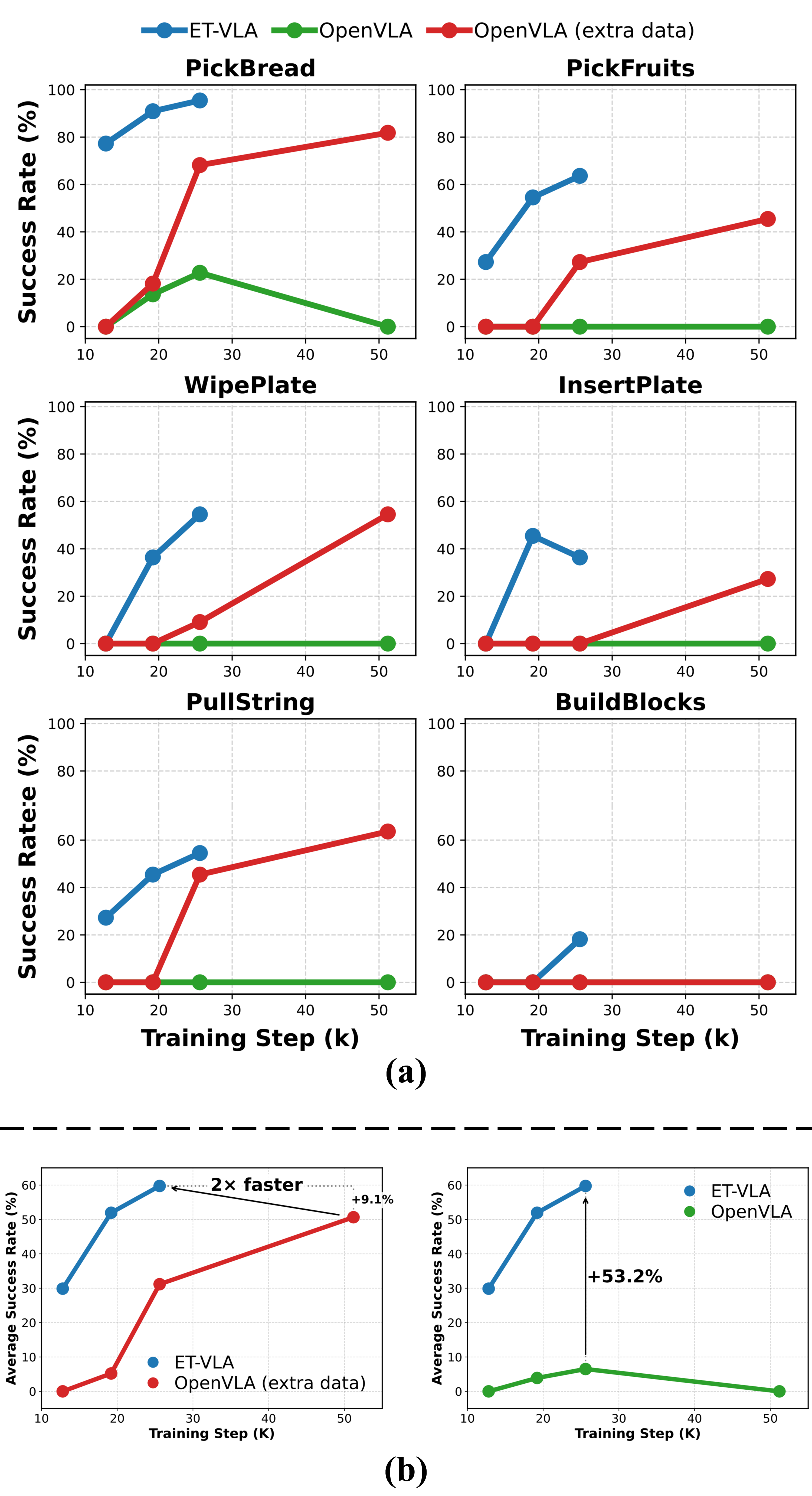}
  \vspace{-0.4cm}
  \caption{\textbf{(a) Learning efficiency.} We show the learning curves of \methodname\space and OpenVLA in 6 real-world tasks. \methodname\space demonstrates a rapid convergence towards high accuracy. \textbf{(b) Success rate over six tasks.} We doubled the train data and extended the training duration by a factor of two for OpenVLA, referring to this result as \textbf{OpenVLA (extra data)}. Under these conditions, ET-VLA outperforms OpenVLA (extra data) by 9.1\% with 2 times less training time.} 
  \vspace{-0.5cm}
  \label{fig:comparison}
\end{figure}

\subsection{Motivation: How does VLA fail on multi-robot}
\label{sec:motivation}
While Vision-Language Action (VLA) models have shown impressive capabilities in learning robot actions and generalizing well in single-arm (unimanual) tasks, it's unclear whether this performance translates directly to more complex systems like multi-robot platforms.  This section investigates this question by evaluating the transferability of VLAs to a bimanual UR5e setup, which serves as a simplified multi-robot system, using the state-of-the-art OpenVLA model. Our physical setup consists of two UR5 arms, each equipped with a wrist-mounted Intel RealSense 435i camera.  An external Intel RealSense 457 camera is positioned in front of the robots. We collected a dataset of 458 trajectories across six different tasks, detailed in Section~\ref{sec:realworld_robot}.  Experiments denoted with "extra data" utilize an additional dataset of 980 human demonstration trajectories. Specifically, we find two major types of failure:

1) The autoregressive nature of VLAs presents a challenge for generating the full set of action tokens required for multi-robot manipulation (seven tokens per robot in our case).  We observed a roughly 50\% failure rate in generating the correct number of tokens, significantly impacting performance. To investigate if insufficient training was the root cause, we doubled the training duration and incorporated the extra human demonstration data. This extended training did enable the model to consistently generate the required number of tokens. We believe this issue arises from limitations in the pre-training data.  Pre-training is crucial for VLAs, enhancing generative capabilities and accelerating downstream task convergence.  OpenVLA, for example, is pre-trained on Open X-Embodiment, a large-scale dataset containing only unimanual robot data. Consequently, the model learns to generate a limited number of action tokens (around six or seven). This unimanual pre-training hinders the model's ability to generate the correct number of tokens when applied to bimanual robots, especially with limited fine-tuning data and iterations.

Even when OpenVLA does generate the correct number of tokens, the average task success rate plateaus on most tasks.  Figure~\ref{fig:comparison}(a) illustrates this for the BuildBlocks task: even with extra data, OpenVLA's success rate remains at 0.  Similar plateaus are visible on other tasks.  Doubling the training data and duration (red line) only marginally improves performance, showing a flat learning curve compared to ET-VLA (blue line).  Figure~\ref{fig:comparison}(b) further demonstrates that simply increasing fine-tuning steps without additional data does not improve OpenVLA's performance. Regarding the average success rate over six tasks, with the same training data and duration, ET-VLA achieves a 53.2\% higher success rate than OpenVLA. While adding extra data to OpenVLA does lead to some improvement, it still underperforms significantly compared to ET-VLA.  Even with double the training time and data, ET-VLA outperforms OpenVLA by approximately 9\%.

2) For tasks requiring multi-robot collaboration, such as opening a bag to place a ball inside, the model frequently skips intermediate actions, proceeding directly to the subsequent task. For instance, the fine-tuned OpenVLA bypasses the initial step of opening the bag and grabs the tennis ball directly. This limitation suggests that the standard autoregressive VLA lacks effective planning capabilities for varied embodiments and task allocation.

Based on these observations and analyses of autoregressive VLA’s multi-robot performance, we propose new methods to enhance autoregressive VLA’s effectiveness in manipulation tasks.

\begin{figure}[t]
    \centering
    \includegraphics[width=0.45\textwidth]{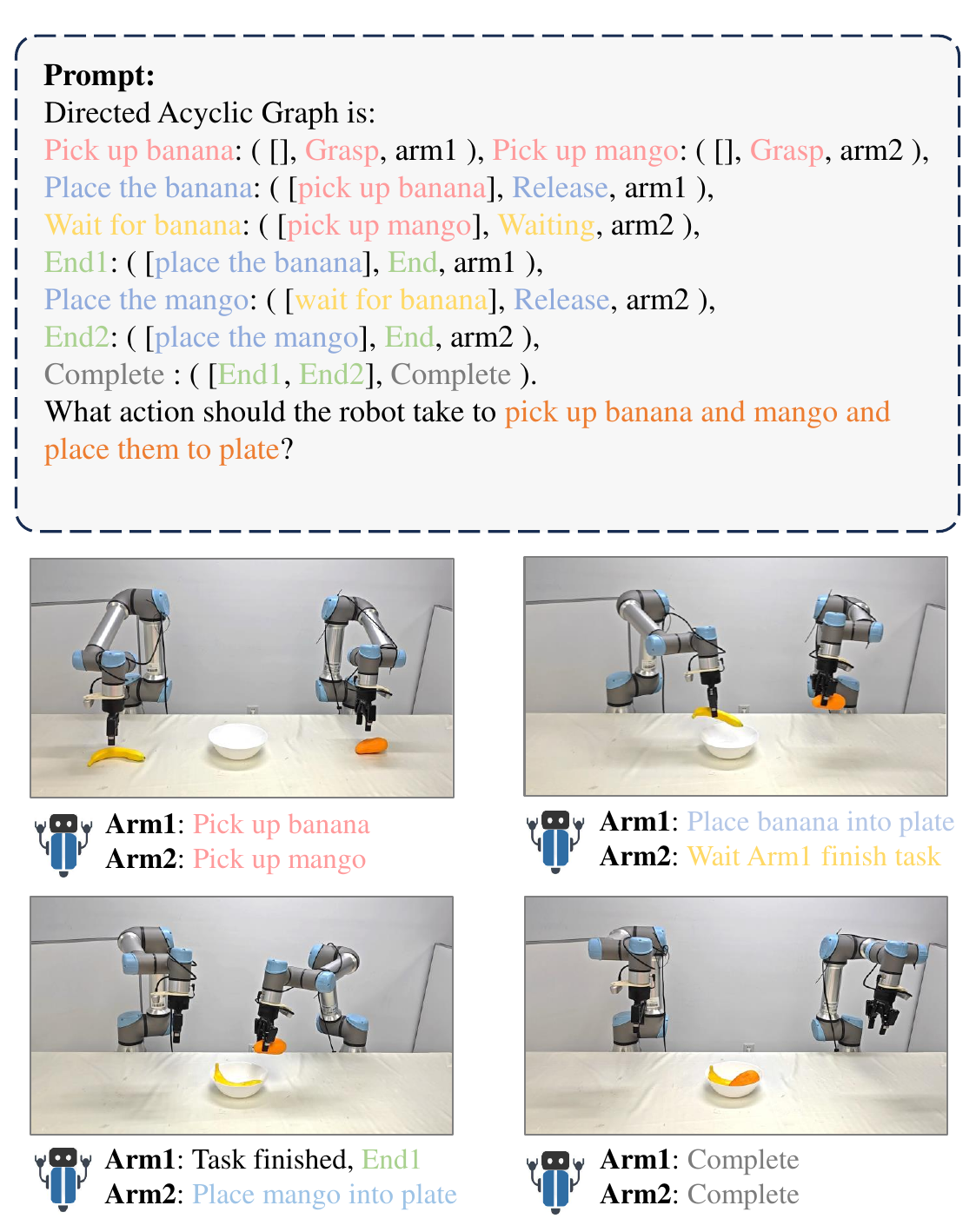}
    \caption{\textbf{Example of Embodied Graph-of-Thought (EGoT)}. To facilitate better understanding for our readers, we provide only the simplified version of the prompt and tasks output. }\label{fig:examples}
    \vspace{-0.4cm}
\end{figure}

\begin{table*}[t]
\centering
\vspace{0.1cm}
\caption{\textbf{Quantitative results in real-world experiments.} We report the average success rate across multiple tasks for all models.}
\label{tab:ur5_exp}
\resizebox{0.95\textwidth}{!}{\begin{tabular}{c|cccccc|c}
\toprule
 &   \multicolumn{6}{c}{Real World Experiment} &  \\
\textbf{Model $\setminus$ Tasks}  & PickBread & PickFruits & WipePlate & InsertPlate & PullString & BuildBlocks & Avg.\\
\midrule
Diffusion Policy~\cite{diffusion-policy} &  6/22 & 4/11 & 5/11 & 2/11 & 5/11 & 0/11 & 22/77  \\ 
TinyVLA~\cite{wen2024tinyvla} & 19/22  & 6/11 & 0/11 & 0/11 & 6/11 & 1/11 & 32/77 \\

OpenVLA~\cite{openvla} &   5/22  &  0/11 & 0/11  & 0/11 & 0/11 & 0/11 & 5/77 \\
\midrule
\textbf{\methodname} & \colorbox{AliceBlue}{\textbf{21/22}} & \colorbox{AliceBlue}{\textbf{7/11}} & \colorbox{AliceBlue}{\textbf{6/11}} & \colorbox{AliceBlue}{\textbf{4/11}} & \colorbox{AliceBlue}{\textbf{6/11}} & \colorbox{AliceBlue}{\textbf{2/11}} & \colorbox{AliceBlue}{\textbf{46/77}} \\ 
\bottomrule
\end{tabular}}
\vspace{-0.1cm}
\end{table*}

\subsection{Synthetic Continued Pretraining}

In the previous section, we identified that autoregressive VLA can fail to generate the required number of action tokens for controlling a multi-robot system due to insufficient training data. A straightforward approach to address this issue would be to train the model with multi-robot data. However, this method presents significant challenges. First, pre-training demands substantial computational resources. For example, OpenVLA was trained on 64 A100 GPUs over 14 days. Second, multi-robot data is scarce, and the size of available bimanual datasets is several orders of magnitude smaller than that of large datasets such as Open X-Embodiment, which contains over 1 million real robot trajectories, the majority of which are unimanual robotic systems. Therefore, it is crucial to explore alternative methods to enable OpenVLA for multi-robot manipulation with limited data. We propose generating synthetic multi-robot data to train a multi-robot-capable autoregressive VLA.

Our focus is to ensure that VLA models predict the correct number of action tokens. For each data sample $x$, where $x = \{\text{observation, instruction}\}$, the standard VLA maps the sample $x$ to a set of 7 action tokens $A = \phi(x)$, with $\text{len}(A) = 7$. To address the limitations noted above, we create synthetic robot data. Specifically, for each sample $x$, we aim for the mapping function to produce 14 action tokens, representing 7 Degrees of Freedom (DoF) for each robot. To generate a valid yet distinct set of tokens for the additional 7 DoF, we introduce a cross-sampling approach. 

Assuming a batch size of $n$ samples $X_{n} = \{x_{1}, x_{2}, \ldots, x_{n}\}$, for a given sample $x_{i}$, we randomly select another sample $x_{j}$, where $i \neq j$. We then append the predicted action tokens $A_{j}$ to $A_{i}$, resulting in a ground truth of 14 action tokens for a sample $x_{i}$. Through this simple operation, the model learns valid actions for robots. Although the image/instruction-to-action mapping may not always be entirely correct, this method ensures that the autoregressively trained VLA learns the token-to-robot mapping. In fact, we used this method to train the VLA model on the BridgeData V2~\cite{walke2023bridgedata} for one epoch. As shown in Table \ref{tab:part_ablation} we demonstrate that this approach is essential for VLA, improving OpenVLA's success rate from 6.49\% to 37.66\%.

\subsection{Embodied Graph-of-Thought}
In Section~\ref{sec:motivation}, we discuss the limitations of existing Vision-Language Action (VLA) models in performing an example of multi-robot tasks, identifying two key challenges. While continued pretraining on synthetic data can address the first challenge, the second, more fundamental issue remains: VLAs struggle with task planning, particularly in coordinating actions between multiple robots. This is a crucial requirement for multi-robot, as parallel task execution necessitates careful synchronization, which linear models typically fail to exploit. Without the ability to properly sequence and coordinate actions, robots are unable to perform complex tasks that require simultaneous or interleaved operations.

To address this limitation, we propose Embodied Graph-of-Thought (EGoT), a novel approach that enables VLAs to understand and manage the temporal dependencies inherent in multi-robot tasks. EGoT decomposes complex tasks into an action graph that explicitly represents these dependencies, providing a structured and interpretable framework for planning. This graph allows the robot to reason about the order in which sub-tasks should be executed, considering the constraints and relationships between them.

This graph better represents the complex temporal dependencies of multi-robot collaboration. We use two robots as an example here, where this framework can easily extend to more robots operating simultaneously. We define the graph as $G = (V, E, T, N )$, where $V$ denotes the tasks, $E$ represents
the dependencies, T categorizes the task types, and $N$ specifies the robot requirements. Each vertex $v_i \in V$ corresponds to a specific sub-task, and each directed edge $e_{ij} = (v_i, v_j ) \in E$ indicates
that $v_i$ must be completed before $v_j$ . The task types include: 1) Grasp, where tasks involve
the engagement of the robot’s gripper and the robot will be occupied; 2) Release, for tasks where an object is released
from the gripper, often associated with placement or release into a specific location; 3) Waiting, where the robot pauses and waits for the other robot to complete its task. This is designed to facilitate coordination and synchronization between the two robots during collaborative operations; 4) End, indicates that all tasks assigned to the robot have been completed, marking the conclusion of its activity in the task sequence; and 5) Complete, serves as the terminal node in the task graph, signifying the successful execution of all tasks and the end of the workflow. This classification aids in specifying the nature
of the task and the number of robots required, thereby enabling a more sophisticated and efficient
planning strategy tailored for multi-robot systems.

\section{Experiments}
In this section, we aim to answer the following question via our experiments:
\begin{itemize}
    \item How does \methodname\space perform on real robots compared to other state-of-the-art VLA models, such as OpenVLA?
    \item How does ET-VLA compare with other non-VLA methods, such as Diffusion Policy?
    \item How important are our proposed techniques — synthetic continued pretraining and embodied graph-of-thought — to overall performance?
\end{itemize}

\begin{table*}[t]
\centering
\caption{\textbf{Performance of different 2D methods on various tasks in RLBench2~\cite{grotz2024peract2} simulation benchmark.} }
\resizebox{\textwidth}{!}{
\begin{tabular}{l|ccccccccccccc|c}
\toprule
Method  & (a) box & (b) ball & (c) buttons & (d) plate & (e) drawer & (f) fridge & (g) handover & (h) laptop & (i) rope & (j) dust & (k) tray & (l) handover easy & (m) oven & Avg.\\
\midrule
ACT~\cite{mobile_aloha} & 0\% & 36\% & 4\% & 0\% & 13\% &0\%  & 0\% & 0\% & 16\% & 0\% & 6\% & 0\% & 2\% & 5.9\%\\
OpenVLA~\cite{openvla} & 0\% & 0\% & 6\% & 0\% &10\% & 0\% & 0\% & 0\% & 0\% & 0\% & 0\% & 0\% & 0\%  &1.2\%\\
\midrule
\textbf{ET-VLA} & \textbf{8\%} & \textbf{56\%} & \textbf{7\%} & \textbf{1\%} & \textbf{19\%} & 0\% & 0\% & \textbf{3\%} & \textbf{21\%} & 0\% & \textbf{11\%} & 0\% & \textbf{6\%} & \textbf{10.2\%}\\
\bottomrule
\end{tabular}
}
\label{tbl:rlbench2}
\vspace{-0.3cm}
\end{table*}
\subsection{Results on Real Robots}
\label{sec:realworld_robot}

\textbf{Implementation Details.} The experimental setup is depicted in Fig~\ref{fig:workspace}. For real-world robotic tasks, we use two UR5 robot arms with Robotiq gripper. We use a single, fixed Realsense D457 camera positioned above the robot. We only use a single RGB image with a resolution of 224 $\times$ 224. Since OpenVLA employs only one extrinsic camera view, we maintained consistency for fair comparison by not using additional wrist cameras. In the synthetic continued pretraining phase, we applied a learning rate of 2e-5, aligning with the rate used in OpenVLA’s pretraining phase, and trained for one epoch. During the fine-tuning stage, we used 50–100 demonstrations for each task and trained all tasks together in a mixed setting. Fine-tuning was performed with an initial learning rate of 2e-4, employing cosine learning rate decay over 20 epochs. All models were trained on 16 A100 GPUs with the same total number of training epochs. To ensure unbiased results, we report only the final checkpoint, avoiding selective reporting.

\noindent
\textbf{Task Description.} 
As illustrated in Fig~\ref{fig:workspace}, we designed six collaborative tasks for our real-world experiment: \textbf{PickBread}, \textbf{PickFruits}, \textbf{WipePlate}, \textbf{InsertPlate}, \textbf{PullString}, and \textbf{BuildBlocks}. Objects were placed randomly within a small range, and we report the average success rate for each method.

1) \textbf{PickBread}: The model chooses which robot to pick up bread from one side of the table and place it on a central plate while keeping the other robot stationary, ensuring robot independence.

2) \textbf{PickFruits}: A banana and a mango on opposite sides of the table are picked up by different robots and placed sequentially in a bowl, testing grip adaptation and task order.

3) \textbf{WipePlate}: One robot holds a sponge, and the other remains stationary as the model cleans a plate with a complex wiping motion, requiring distinct robot coordination.

4) \textbf{InsertPlate}: A plate is passed between robots to be inserted into a rack with precise end-effector (EEF) posture control, testing fine motor skills.

5) \textbf{PullString}: The model grasps and aligns both ends of a string marked with tape, ignoring its initial shape, and pulls it straight to a marked spot.

6) \textbf{BuildBlocks}: robots construct a block house by stacking cylinders and placing a wedge as a roof, demanding precision and error-free sequencing.

These tasks evaluate the model's ability to handle independent and coordinated robot movements across varied scenarios.

\noindent
\textbf{Real-world experimental results.} 
We evaluated our method against Diffusion Policy (DP)~\cite{diffusion-policy}, TinyVLA~\cite{wen2024tinyvla} and OpenVLA~\cite{openvla}. For a fair comparison, we used only a single RGB image with a resolution of 224 × 224 to train all baselines, as OpenVLA employs a single extrinsic camera view. Furthermore, our approach integrates language instructions throughout the process, aligning with the input used in all the baselines.

The experimental results are shown in Table~\ref{tab:ur5_exp}. \methodname\space achieved the highest success rate across all tasks, surpassing other baselines. Notably, the InsertPlate and BuildBlocks tasks require the model to control the end-effectors to perform relatively complex rotational operations. Additionally, TinyVLA and Diffusion Policy require extra state data to predict actions, whereas \methodname \space relies solely on input images. We also observed that OpenVLA exhibits an extremely low success rate, achieving 0\% on five tasks. These results reinforce our earlier observation that autoregressive VLAs, even when trained on large-scale, multi-embodied data, may not effectively transfer their capabilities to multi-robot. While our method, ET-VLA, builds upon OpenVLA, the significant performance improvement clearly demonstrates the importance of our proposed techniques.

\begin{figure}[t]
    \centering
    \vspace{0.2cm}
    \includegraphics[width=0.4\textwidth]{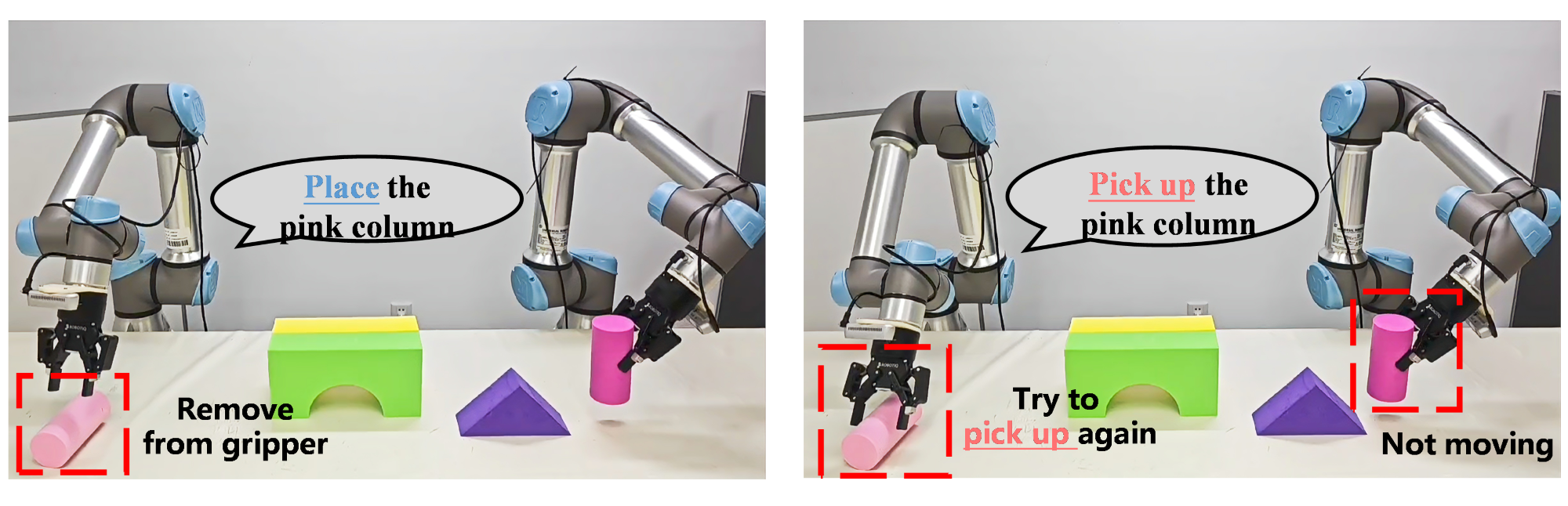}
    \vspace{-0.2cm}
    \caption{\textbf{EGoT is capable of handling complex task sequences.} We deliberately remove the column from the gripper and place it back on the table. we observe the robot try to pick up again. And the model’s output transitions to ”pick up the pink column”.}\label{fig:block}
    \vspace{-0.4cm}
\end{figure}

\begin{figure}[h]
    \centering
    \includegraphics[width=0.35\textwidth]{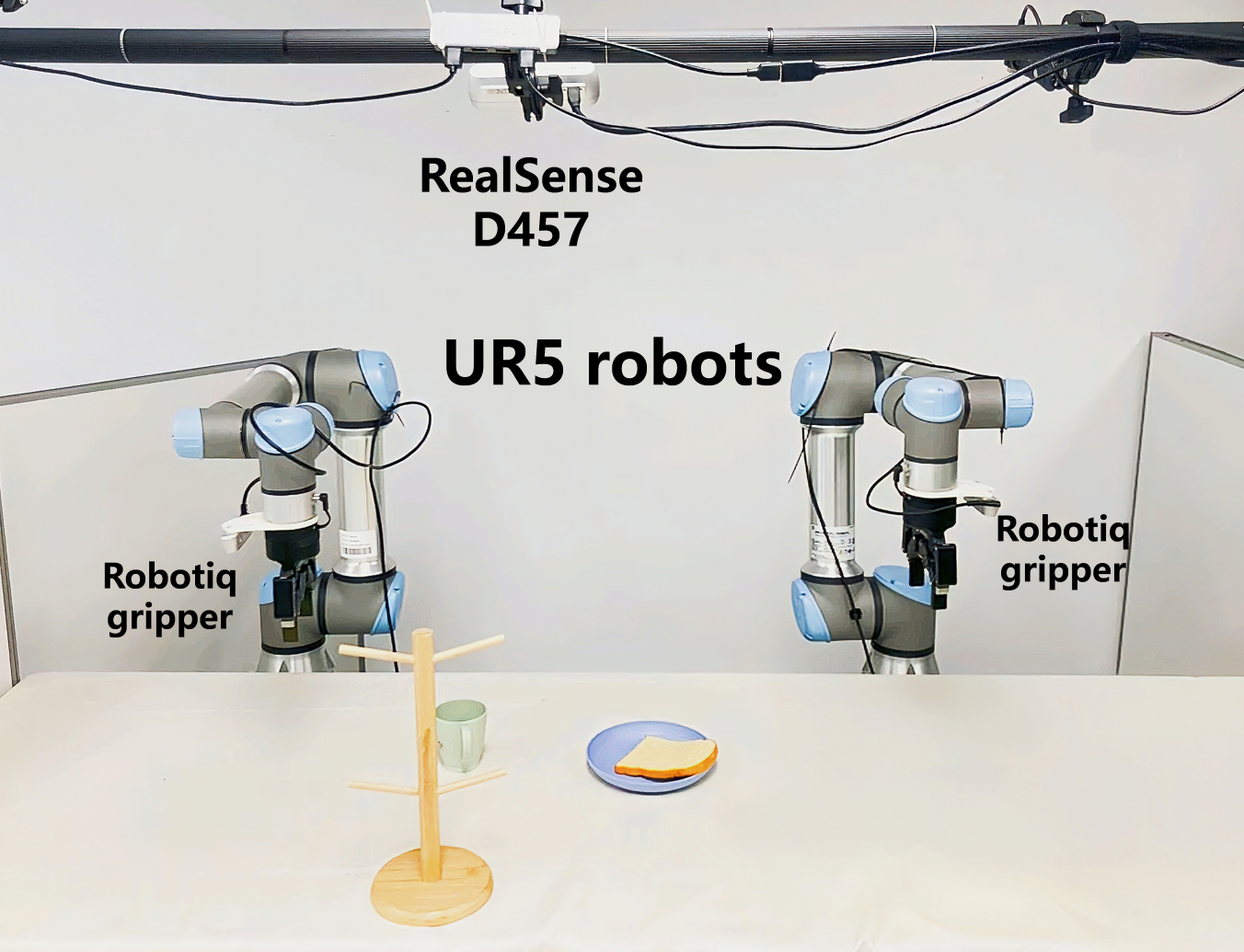}
    \vspace{-0.3cm}
    \caption{\textbf{Real-world experiment setup.} }
    \label{fig:setup}
\end{figure}

\noindent
\textbf{Real-world experiment setup} 
In our real-world experiment setup, we use two UR5 robot arms equipped with Robotiq grippers, as is shown in Figure \ref{fig:setup}. A single, fixed RealSense D457 camera is positioned overhead to capture visual observations. We rely solely on a single RGB image with a resolution of 224 × 224. Since OpenVLA utilizes only one extrinsic camera view, we ensure a fair comparison by not incorporating additional wrist cameras.

\begin{table}[t]
\vspace{-0.2cm}
\centering
\caption{\textbf{Experiment on RoboTwin simulation~\cite{mu2024robotwin}}. We compare the Diffusion Policy using 50 demonstrations.}
\resizebox{0.48\textwidth}{!}{\begin{tabular}{l|c|c|c}
\toprule
Task & Diffusion Policy & OpenVLA & \textbf{ET-VLA}   \\
\midrule
Apple Cabinet Storage & 72\% & 16\% &  \textbf{85\%}  \\
Block Handover & 28\% & 7\%  &\textbf{45\%} \\
Blocks Stack (Easy) & 2\% & 0\% &\textbf{18\%} \\
Container Place & 0\% & 0\% &\textbf{15\%}  \\
Dual Bottles Pick (Easy) & 54\% & 10\% & \textbf{70\%} \\
Empty Cup Place & 20\% & 0\% & \textbf{35\%}  \\
Shoe Place & 9\% & 0\% & \textbf{24\%} \\
Block Hammer Beat & 0\% & 0\% & \textbf{12\%}  \\
Block Sweep & 95\% &  24\% & \textbf{96\%}  \\
Blocks Stack (Hard) & 0\% & 0\% & \textbf{8\%}  \\
Diverse Bottles Pick & 0\% & 0\% & \textbf{14\%}  \\
Dual Bottles Pick (Hard) & 28\% & 0\& & \textbf{42\%}  \\
Mug Hanging & 0\% & 0\% &\textbf{10\%}  \\
Shoes Place & 0\% & 0\%& \textbf{13\%}  \\
\midrule
\textbf{Average} & 27.7\% & 4.1\% & \textbf{40.1\%} \\ 
\bottomrule
\end{tabular}}
\label{tab:robotwin}
\vspace{-0.7cm}
\end{table}
\begin{table*}[t]
\centering
\vspace{-0.1cm}
\caption{
\textbf{Ablation study for two components in \methodname.} We report the average success rate across multiple tasks.}
\label{tab:part_ablation}
\resizebox{\textwidth}{!}{\begin{tabular}{c|cccccc|c}
\toprule
 &   \multicolumn{6}{c}{Real World Experiment} &  \\
\textbf{Model $\setminus$ Tasks}  & PickBread & PickFruits & WipePlate & InsertPlate & PullString & BuildBlocks & Avg.\\
\midrule
\textbf{\methodname} & \textbf{21/22} & \textbf{7/11} & \textbf{6/11} & \textbf{4/11} & \textbf{6/11} & \textbf{2/11} & \textbf{46/77} (59.74\%) \\ 
- Embodied Graph-of-Thought & 17/22 & 5/11 & 2/11 & 1/11 & 4/11 & 0/11 & 29/77 (37.66\%) \\
- Synthetic Continued Pretraining  & 5/22  &  0/11 & 0/11 & 0/11  & 0/11 & 0/11  & 5/77 (6.49\%)\\ 
\bottomrule
\end{tabular}}
\vspace{-0.5cm}
\end{table*}

\subsection{EGoT is Real-Time Action Interpreter}
Our previous section demonstrates that Embodied-GoT (EGoT) effectively improves model performance and generalization. In this section, we show that, in addition to enhancing the model’s ability to learn robotic actions, EGoT also provides interpretable language, enabling users to understand the model’s current "thought" process.

As shown in Fig. \ref{fig:block}, EGoT is capable of handling complex task sequences. For example, in the BuildBlocks task, EGoT follows a task sequence where the left robot picks up the pink column while the right robot pick up another. After the pink column is picked up, the model's output transitions to "place the pink column". If we deliberately remove the column from the left robot and place it back on the table, we observe the robot hesitating and try to pick up the column again. And the model's output transitions to "pick up the pink column". For a typical observer, it may be unclear whether the model is processing this new state or abandoning the task. However, with our approach, EGoT updates to reflect the new state, such as "pick up the pink column", giving observers clear insight into the model's current focus without ambiguity. This interpretability is an emergent property of EGoT and does not require specific training. 

\subsection{Simulation Experiments}

We further conduct experiments on two simulation benchmarks, the RLBench2~\cite{grotz2024peract2} and RoboTwin~\cite{mu2024robotwin}.

\textbf{RLBench2.} RLBench2 extends the original RLBench benchmark~\cite{james2020rlbench} from unimanual to bimanual robot manipulation.  This new benchmark comprises 13 core bimanual tasks, totaling 23 unique task variations.  RLBench2 retains the core functionality and key properties of RLBench.  Standard baselines, including ACT~\cite{mobile_aloha}, RVT~\cite{goyal2023rvt}, and PerACT~\cite{shridhar2023perceiver}, are provided. However, as our approach uses only 2D input, we compare solely against ACT, since RVT and PerACT require 3D input.  Additionally, we implemented OpenVLA~\cite{openvla} using the publicly available model and pre-trained weights. We also initialized our model with these pre-trained OpenVLA weights.

As demonstrated in Table~\ref{tbl:rlbench2}, ET-VLA significantly outperforms ACT and OpenVLA in manipulation tasks, achieving a 10.2\% average success rate compared to 5.9\% for ACT and 1.2\% for OpenVLA. While ET-VLA excels in tasks like "ball" and "rope," performance varies considerably across tasks for all methods, with challenges remaining in areas like "fridge" and "oven." OpenVLA demonstrates consistently low performance, while ACT shows moderate success. These results highlight the need for our approach to improve the robustness and generalization of beyond unimanual manipulation methods.

\textbf{RoboTwin.} We assess the performance of our method on RoboTwin~\cite{mu2024robotwin}, a new simulation benchmark specifically designed for bimanual robot manipulation using AgileX robots.  This benchmark includes a variety of tasks. We compare our approach to the Diffusion Policy~\cite{diffusion-policy}, a well-established baseline for visuomotor policy learning.  For all experiments, we used a standard image resolution of $320 \times 240$ for the camera input.

Table~\ref{tab:robotwin} presents a comparative performance analysis between Diffusion Policy and ET-VLA across 14 distinct robot manipulation tasks. The data clearly demonstrates the superior performance of ET-VLA, achieving a significantly higher average success rate of 40.1\% compared to the Diffusion Policy's 27.7\%.  Across nearly all individual tasks, ET-VLA exhibits a notable improvement in performance. For instance, ET-VLA achieves an 85\% success rate in "Apple Cabinet Storage" compared to Diffusion Policy's 72\%, and a 45\% success rate in "Block Handover" where Diffusion Policy only scores 28\%.  Even in challenging tasks like "Blocks Stack (Easy)" and "Container Place," where Diffusion Policy struggles with very low success rates, ET-VLA achieves significantly higher, albeit still moderate, performance.  This consistent outperformance across the majority of tasks highlights the effectiveness of ET-VLA as a solution for robot manipulation challenges.

\subsection{Ablation Study}
\label{sec:ablation}

To evaluate the contributions of the proposed SCP and EGoT components in our framework, we conducted two ablation experiments: 1) Removing EGoT while retaining SCP. 2) Removing both EGoT and SCP, leaving only the baseline OpenVLA.
These experiments aim to analyze the performance degradation caused by the absence of these components, demonstrating their individual and combined impacts on the overall success rates. The experimental results are demonstrated in Table~\ref{tab:part_ablation}.

In the first ablation experiment, the EGoT module was removed, while the SCP component was retained. This setup evaluates the contribution of EGoT in enhancing task planning capabilities and its awareness of task sequencing for different robots. The model's performance dropped significantly, particularly on tasks such as WashPlate and PlacePlate, which rely heavily on the task planning capabilities provided by EGoT. In contrast, the performance on the PickBread task, which requires only one robot, showed only a slight decline.

In the second ablation experiment, both the EGoT and SCP components were removed, leaving only the baseline OpenVLA. This setup evaluates the combined impact of these two modules on task success. As expected, the performance dropped further compared to the first experiment, with success rates significantly lower across all tasks.  Although the PickBread task requires only one robot, it demands the model to generate all 14 action tokens accurately. However, the model occasionally failed to produce the full sequence of 14 tokens and often moved the arm aimlessly, resulting in task failures.

\section{Conclusion}
Vision-language-action (VLA) models are widely used for developing robust and generalizable robotic policies. However, current VLA models are not designed for multi-robot applications. In this work, we empirically analyze the failure modes of VLA models on multi-robot manipulation tasks. Specifically, we identify two key issues causing failure: 1) VLA models often fail to generate a sufficient number of action tokens to control multi-robot, and 2) VLA models struggle to execute correct action sequences for parallel robots. To address these challenges, we propose synthetic continued pretraining to enable VLA models to produce the correct number of action tokens, followed by an Embodied Graph-of-Thought approach to enhance the model's awareness of multi-robot task sequences. Our approach is validated through real-world robotic tasks, where we demonstrate that our methods improve the average success rate of state-of-the-art VLA models on real robots from  6.49\% to 59.74\%.

\nocite{langley00}

\bibliography{bib_files/robot, bib_files/vlm, bib_files/main}
\bibliographystyle{icml2025}


\newpage






\end{document}